# Federated Learning for Clinical Structured Data: A Benchmark Comparison of Engineering and Statistical Approaches

Siqi Li[1#], Di Miao[1#], Qiming Wu[1], Chuan Hong[2], Danny D'Agostino[1], Xin Li[1], Yilin Ning[1], Yuqing Shang[1], Huazhu Fu[3], Marcus Eng Hock Ong[4,5,6], Hamed Haddadi[7], Nan Liu[1,4,8]*

[1] Centre for Quantitative Medicine, Duke-NUS Medical School, Singapore, Singapore

[2] Department of Biostatistics and Bioinformatics, Duke University, Durham, NC, USA

[3] Institute of High Performance Computing, Agency for Science, Technology and Research, Singapore, Singapore

[4] Programme in Health Services and Systems Research, Duke-NUS Medical School, Singapore, Singapore

[5] Health Services Research Centre, Singapore Health Services, Singapore, Singapore

[6] Department of Emergency Medicine, Singapore General Hospital, Singapore, Singapore

[7] Department of Computing, Imperial College London, London, England, United Kingdom

[8] Institute of Data Science, National University of Singapore, Singapore, Singapore

[#] Co-first author

* Correspondence: Nan Liu, Centre for Quantitative Medicine, Duke-NUS Medical School, 8 College Road, Singapore 169857, Singapore. Phone: +65 6601 6503. Email: liu.nan@duke-nus.edu.sg

**Abstract**

Federated learning (FL) has shown promising potential in safeguarding data privacy in healthcare collaborations. While the term "FL" was originally coined by the engineering community, the statistical field has also explored similar privacy-preserving algorithms. Statistical FL algorithms, however, remain considerably less recognized than their engineering counterparts. Our goal was to bridge the gap by presenting the first comprehensive comparison of FL frameworks from both engineering and statistical domains. We evaluated five FL frameworks using both simulated and real-world data. The results indicate that statistical FL algorithms yield less biased point estimates for model coefficients and offer convenient confidence interval estimations. In contrast, engineering-based methods tend to generate more accurate predictions, sometimes surpassing central pooled and statistical FL models. This study underscores the relative strengths and weaknesses of both types of methods, emphasizing the need for increased awareness and their integration in future FL applications.

# INTRODUCTION

Privacy regulations, such as the European Union General Data Protection Regulation[1], have introduced significant challenges to traditional data-sharing strategies in cross-institutional medical research collaborations. Consequently, federated learning (FL) has emerged as a trending solution[2] to address data privacy concerns in the healthcare industry, enabling research collaborations without the need for data sharing[3]. FL is a machine learning (ML) paradigm that enables multiple participating sites, referred to as clients, to collaboratively solve a modeling problem without the need to exchange or transfer data[4]. Since its adoption in healthcare, FL has enhanced medical research not only by expanding the scope of research partnerships, but also by developing and implementing robust models[3]. For example, during the COVID-19 pandemic, Dayan et al.[5] conducted a study in which they constructed clinical outcome prediction models that outperformed local models by federating data from 20 institutes across the globe.

In the context of clinical FL, an aspect that is often overlooked but is equally critical as prediction tasks, is the need to accurately estimate the association between important factors and clinical outcomes, a concept known as "point estimation", which plays a vital role in guiding the development of interventions and resource allocation strategies. Much like prediction tasks, point estimation tasks can also benefit from FL frameworks by leveraging information from external sources while addressing data privacy concerns. These approaches aim to mitigate bias in point estimates, deviating from the traditional emphasis on optimizing predictive power. For instance, Duan et al.[6] assessed the relative bias of estimated coefficients through simulation studies, and employed real-world data to showcase their FL algorithm's comparable performance to pooled analysis when considering estimated odds ratios and their confidence intervals (CIs) for medications associated with fetal loss.

While the engineering community formally introduced the term "FL"[7], the statistical field had been investigating similar privacy-preserving algorithms under different names, such as "distributed learning"[8,9] and "distributed algorithms"[10,11]. However, compared to engineering-based methods, statistics-based FL algorithms have not gained much attention[12] in healthcare research, where ensuring privacy can be of utmost importance.

The fundamental distinction between engineering-based and statistics-based FL algorithms lies in their model agnosticism[12] – whether they can be applied across various types models or are specific to a particular type. FL algorithms originating from the engineering community typically prioritize predictive power and usually develop model-agnostic FL frameworks, making them versatile for use with different statistical or ML models, including, but not limited to, traditional regression models and various types of neural networks. In contrast, statistics-based methods tend to place greater emphasis on the accuracy of point estimation and, as a result, they are typically developed with a model-specific focus, tailored to meet the decentralized requirements of a single statistical model. This specialization limits their adaptation to other models.

Another distinction lies in the fact that statistics-based methods often prioritize the fundamental task of statistical inference, while engineering-based methods may be primarily adapted for prediction tasks. While prediction tasks have become the predominant focus across the various fields employing ML, it is crucial to recognize that healthcare and clinical science also place importance on traditional studies involving non-prediction tasks, especially when using structured data. Beyond predictions, a variety of tasks also require execution within privacy-preserving frameworks. Notable examples include exploring connections between exposures and outcomes[6,13], phenotyping[14,15], and even optimizing individualized treatment[16], all of which utilize model-specific FL methods. Since these tasks involve non-predictive aspects and often require statistical inferences, model-agnostic FL algorithms may not be well-suited for these purposes.

Consequently, there is a need for further exploration to understand how different types of healthcare research centered on structured data may benefit from distinct advantages offered by statistics-based and engineering-based privacy-preserving algorithms[12]. To our knowledge, no empirical comparisons between FL methods from these two distinct fields have been reported. This benchmarking study aims to bridge this gap by evaluating FL frameworks from both engineering and statistical domains. Specifically, we apply various FL frameworks on both simulated and real-world data, aiming to evaluate the performance in terms of measuring bias and uncertainty of point estimates, as well as prediction accuracy. We also seek to establish a comprehensive tutorial and recommendations for future researchers.

## RESULTS
**Overview of FL algorithms and data**
We evaluated five FL frameworks, including GLORE[17], FedAvg[7], FedAvgM[18], $q$-FedAvg[19], and FedProx[20]. Among them, GLORE, FedAvg and FedProx have been commonly employed[12] in previous FL studies involving clinical structured data. The other two, FedAvgM and $q$-FedAvg, have had relatively limited[12] application in clinical structured data so far but have drawn our attention. Our evaluation encompassed both simulated and real data with binary outcomes.
We conducted simulations using data from three client sites referred to as 'Site1', 'Site2' and 'Site3', along with a central dataset created by aggregating data from all clients, which we will refer to as 'central.' Our simulations covered three primary scenarios: data distribution shifts based on mean, data distribution shifts based on variance, and model shifts, as summarized in

Table 1. In each of these simulation scenarios, we conducted two experiments: one with a small sample size and another with a large sample size, aligning with the respective shifting patterns. In addition to using simulated data, we formed cohorts from two real-world electronic health records (EHR) datasets: MIMIC-IV-ED[21] and the Singapore General Hospital (SGH) emergency department (ED) data[22]. We conducted federations separately for homogeneously and heterogeneously partitioned SGH and MIMIC data. Consistent with the simulated data configuration, we established an equal number of clients for artificially partitioned datasets. Furthermore, we conducted a federation between the MIMIC and SGH datasets, resulting in two clients. More specific information about the cohort formation settings is available in Table 1.

**Performance of prediction tasks**
The performance of binary outcome prediction is assessed using the receiver operating characteristic (ROC) curve analysis. Figure 1 illustrates model comparisons within the context of Setting I in the simulation studies, which introduces a shift of covariate means by 10% and 20% while working with a relatively small sample size. The subplots are organized vertically to represent the degree of shifting and horizontally to showcase different testing datasets. Figure 1 reveals two key observations: 1) as the sample size increases from site 1 to site 3, the overall performance of all models improves; 2) within a single site, no significant differences are observed across all five FL frameworks in terms of prediction task accuracy. Additionally, different choices of parameter 'μ' (proximal term for addressing statistical heterogeneity[20]) for FedProx[20] have minimal impact on the performance of the prediction task. For a more comprehensive view of shifting trends across all settings, please refer to eFigure 1-6 in Section E.1 of the Supplementary Material, which shows similar patterns.

Table 2 summarizes the performance of the prediction task using real data in each setting. An intriguing discovery is that while GLORE produces results closely mirroring those of central analysis, with coefficients that exhibit remarkable similarity (as demonstrated in eTable 2), engineering-based FL models occasionally achieve higher prediction accuracy than centralized analysis. For instance, on the MIMIC data, both the central model and the GLORE model achieve an area under the ROC curve (AUC) value of 0.789. However, the $q$-FedAvg model outperforms all models on MIMIC, with an AUC value of 0.811. It is important to note that such superior performance is not consistent across both datasets. For the SGH data, the $q$-FedAvg model achieves an AUC of 0.842, which is lower than the central model's AUC of 0.850.

**Relative bias of coefficient estimates and confidence intervals (CIs)**
We use violin plots to visualize the relative bias of coefficient estimates by comparing the model-estimated coefficients to the ground truth values, using 100 simulation runs. As shown in Figure 2 (representing Setting I with a relatively small sample size), among GLORE, FedAvg, FedAvgM and $q$-FedAvg, no significant differences are observed in terms of the relative biases of point estimations deviating from the ground truth. However, it is worth noting that the choice of parameter 'µ' for FedProx results in significantly different levels of bias for point estimations. For a comprehensive view of bias in point estimates across all settings, please refer to eFigure 7-12 in Section E.2 of the Supplementary Material, where similar patterns can be observed.

In addition to direct point estimates, it is important to highlight that only GLORE has the capability to estimate confidence intervals (CIs). We report that the estimated probability of coverage consistently exceeds 90% when using simulated data. For more detailed information, including the average lower and upper bounds of CIs, please refer to eTable 3-5 in Section C of the Supplementary Material. However, as expected, when effect sizes exhibit heterogeneity across sites, GLORE may no longer provide reliable coverage for point estimates at each site. Since the ground truth model (i.e., the conditional distribution $Y|X$, where $Y$ is the outcome and $X$ represents all predictors) is unknown with real data, evaluating the bias of estimated parameters for logistic regression is not feasible.

**Communication cost**

The communication cost of all five benchmarked methods on simulated data are detailed in eTable 6. In the latter half of the table, it becomes evident that GLORE demonstrates greater communication efficiency when compared to the four engineering-based FL methods. GLORE consistently required fewer than 6 rounds of communication on average, while all other methods necessitated at least 10 rounds for convergence. The communication cost for real data can be found in eTable 7. Consistent with the findings observed in the simulation studies, GLORE continues to outperform other methods in terms of communication cost when applied to real data. In parallel, when dealing with real data, FedProx exhibits relatively lower communication efficiency, requiring more rounds of communication for convergence compared to the other four methods.

## DISCUSSION

Many studies[23–26] have predominantly focused on evaluating FL approaches within the domain of prediction, while non-prediction tasks such as the estimation of the effects of various factors on outcomes are also important in clinical settings. Lack of guidance for selecting appropriate FL methods for diverse clinical applications underscores the need to discern the varying capabilities of FL methods[12]. We filled this gap by providing practical recommendations for applying FL frameworks to the analysis of clinical structured data, informed by empirical evidence derived from our benchmark study. As illustrated in Figure 3, we comprehensively discuss and offer suggestions regarding FL frameworks for future clinical research, from perspectives of prediction and non-prediction tasks, data heterogeneity, and real-world implementation challenges.

As highlighted by Li et al.[12], in FL studies primarily concerning unstructured data, prediction performance is often employed as the primary and, most of the time, the sole metric for evaluating the success of FL. However, when dealing with structured data, especially those commonly used in traditional clinical and medical studies, the range of tasks is considerably more diverse compared to those in traditional engineering field. In line with Li et al.[12], we adopt the same categorization to classify clinical decision-making tasks into two major categories: prediction and non-prediction tasks. In essence, if the objective is solely to predict an outcome and involves utilizing prediction metrics such as accuracy or those derivable from a confusion matrix, the task falls within the realm of prediction tasks. Conversely, if the goal extends beyond prediction and encompasses tasks such as association studies and phenotyping, it is classified as a non-prediction task.

In the context of non-prediction FL tasks, statistics-based FL frameworks are more suitable than engineering-based approaches in two aspects. Firstly, as shown in eTable 3-5, statistics-based FL enables convenient estimations of CIs for model parameters, providing users with a straightforward insight into the level of confidence associated with these estimates. In this study, only the GLORE framework allows for the calculation of CIs for estimated coefficients in logistic regressions, as detailed in the results section. In contrast, engineering-based techniques necessitate additional development to achieve this capability, primarily by resorting to bootstrap, since model-agnostic frameworks do not inherently offer analytical solutions. Nevertheless, it is worth noting that bootstrap can be computationally intensive, which may account for the absence of currently available implementations of engineering methods featuring bootstrap as an option, to the best of our knowledge.

The second reason is rooted in the theoretical soundness of statistics-based methods, often resulting in more accurate parameter estimations and robust hypothesis testing. This is supported by results in eTable 2, where estimations of coefficients by engineering-based methods exhibit more substantial deviations from the centralized model when compared to the GLORE model. Similar patterns are evident in results from simulated data, as illustrated in Figure 2 and eFigure 7-12, where it becomes apparent that variations in hyperparameters can significantly affect the bias in engineering-based methods. It is worth noting that hyperparameter tuning in engineering-based methods primarily relies on prediction accuracy metrics such as mean squared error (MSE), as in this study. Consequently, selecting the optimal hyperparameters can be a challenging task, as evident in Figure 1 and eFigure 1-6, where the choice of hyperparameters had minimal impact on prediction performance. In light of this empirical evidence, engineering-based methods may introduce more bias than statistics-based methods in non-prediction tasks. Furthermore, these biases may remain undetected in real-world data analysis, where specifying ground truth can be particularly difficult.

Statistics-based and engineering-based FL methods also differ in their approaches to handling data heterogeneity. In the engineering community, statistical heterogeneity in FL is broadly defined as scenarios where the data are not independently and identically distributed (i.i.d.)[27,28]. However, the statistical literature usually distinguishes between heterogeneity in models (conditional distribution of $Y|X$) and heterogeneity in covariate distributions ($P(X)$), recognizing their distinct impacts on model building and inference[29–31]. A notable example of model heterogeneity can be found in the work by Liu et al[31], where they propose debiasing distributed least absolute shrinkage and selection operator[32] (LASSO) capable of handling both model and covariate heterogeneity. Another example of handling model heterogeneity is available in Gu et al.[33], where they proposed a generalized linear model allowing for population-specific intercepts and $X$ coefficients. While these examples contribute to a more thorough understanding of FL data analysis, the complexity of model-specific developments also becomes a significant limitation for statistics-based FL methods, making them more challenging to generalize to different models. For instance, GLORE is limited to handling logistic regression, while engineering based FL solutions can be readily applied to a diverse range of models and clinical research questions[12].

Given the complexity of real-world data, determining the suitability of classic i.i.d.-based FL frameworks without substantial empirical evidence is challenging. Among the five FL frameworks benchmarked in this study, both FedAvg and GLORE were theoretically designed only for i.i.d. data. However, both succeeded in handling both heterogeneous simulated and real data in our experiments. Therefore, future researchers may consider benchmarking classic FL frameworks, which assume i.i.d. scenarios, using heterogeneous datasets to further evaluate their effectiveness. However, this strategy may not always yield optimal results, particularly when dealing with 'partially-Blackbox' heterogeneity (where data heterogeneity is evident, but model heterogeneity is difficult to specify). As a result, ongoing investigations and evaluations are necessary to thoroughly assess the strengths and weaknesses of different FL methods for handling both data and model heterogeneity.

The engineering-based methods applied to real data in this study have, on occasion, exhibited superior prediction performance, outperforming both the central pooled and local models, as evidenced in Table 2. This pattern consistently remained in both the artificially partitioned SGH/MIMIC data and the federation of MIMIC and SGH data. To comprehend this phenomenon, let us explore the optimization strategies. GLORE employs the Newton-Raphson iteration[17] for FL training, involving second-order approximation[34] of the log-likelihood function by using the Hessian matrix. In contrast, FL frameworks based on FedAvg optimize the target loss function through stochastic gradient descent (SGD). Theoretically, SGD exhibits a slower convergence rate compared to the Newton method, with the latter achieving superior accuracy in reaching the optimal solution[35]. As discussed earlier, in the context of non-prediction tasks, this behavior of SGD can be considered disadvantageous when compared to model-specific solutions like GLORE, as it may introduce more bias to parameter estimations. However, when the focus shifts to predictive aspects, this behavior appears advantageous, potentially leading to more generalizable prediction models with superior testing accuracy.

In real world applications, the implementation of statistics-based FL frameworks and engineering-based FL frameworks can present a significantly different level of technical difficulty. For statistics-based methods, iterations usually involve only a few rounds and do not require a central server. FL collaborations, in this case, can be conducted easily as long as participants receive and broadcast informative summary-level statistics to each other. However, for engineering-based methods, which often require at least one central server capable of computation, establishing a secure system with data owners may not be as straightforward as one might imagine. For example, if hospital data can only be accessed via a protected computing system[36] which does not allow connection to outside servers, statistics-based FL methods are more adaptable due to their minimal requirement for the data systems of participating clients.

In summary, both engineering-based methods and statistics-based FL methods come with their own set of advantages and disadvantages. Our study could provide some valuable empirical insights for future researchers to reference when selecting and adopting these methods. A promising direction for future research involves exploring the fusion of engineering-based and statistics-based FL algorithms to enhance engineering-based methods with statistical inference capabilities while increasing the adaptability of statistics-based methods across a wide range of model types.

## METHODS

We selected five representative FL frameworks, which are GLORE[17], FedAvg[7], FedAvgM[18], $q$-FedAvg[19], and FedProx[20]. A brief introduction to these frameworks is available in Table 3, and detailed technical information can be found in Section A of the Supplementary Material. Our evaluation consisted of two main phases: first, we assessed the frameworks using simulated data with known ground truth effect sizes for covariates, providing a controlled testing environment; second, we utilized real-world clinical data to evaluate performance of these frameworks in practical, real-world scenarios, particularly for downstream prediction tasks.

**Simulated data**

Data simulation was conducted using R 4.2.1. We considered a total of three participating sites which provide a realistic representation for cross-silo FL settings in healthcare, often involving multiple institutions or hospitals. Let $p$ denote the total number of predictors, and $s$ denote the number of predictors with non-zero effect sizes. In all simulation settings, we fixed $p = 20$ and $s = 6$. The predictors with non-zero effect size were denoted by $X_i$, where $i = 1, 2, 3, \ldots 6$, and their corresponding non-zero effect sizes were denoted as $\beta_i$, where $i = 1, 2, 3, \ldots 6$. The simulations covered three distinct settings detailed in Table 1: setting I involved covariate mean shifts, setting II addressed covariate standard deviation shifts, and setting III focused on shifts in effect sizes.

The values of $\beta_i$ were set as follows: -2, 1, 0.8, 0.4, 0.2 and 0.1 for $i = 1, 2, 3, \ldots 6$ throughout settings I and II. In setting III, which accounts for shifts in effect size, the values of $\beta_i$ were adjusted to $(1 - \alpha)\beta_i$, $\beta_i$ and $(1 + \alpha)\beta_i$ for site 1, site 2 and site 3, respectively, where α serves as the shifting parameter. Similarly, for setting I and II, corresponding to covariate distribution shifting, the means of covariates were modified to $(1 - \alpha)\mu_i$, $\mu_i$ and $(1 + \alpha)\mu_i$ for site 1, site 2 and site 3, and the standard deviations of covariates were modified to $(1 - \alpha)\sigma_i$, $\sigma_i$ and $(1 + \alpha)\sigma_i$ for site 1, site 2 and site 3.

We conducted all three simulation settings for two distinct sample size scenarios: one with relatively small sample sizes, where the sample sizes for site 1, site 2 and site 3 were 1000, 2000, and 4000, respectively; and another with relatively large sample sizes, where the sample sizes for site 1, site 2 and site 3 were 3000, 6000, and 12,000, respectively. The training and testing datasets were partitioned in a 70% to 30% proportion.

**Real-world datasets**

We utilized two real ED datasets, MIMIC-IV-ED[21] and EHR from SGH for our experiments with real-world data. The EHR data of SGH was extracted from the SingHealth Electronic Health Intelligence System, and a waiver of consent was granted for EHR data collection and retrospective analysis. The study has been approved by the Singapore Health Services' Centralized Institutional Review Board, with all data deidentified.

MIMIC-IV-ED is an open source dataset and we follow the data extraction pipelines by Xie et al.[37] This process resulted in the creation of a master dataset, based on which we performed the following cohort formation procedures. Specifically, we formed a cohort of 9071 samples by filtering the master dataset to include only ED admissions of Asian patients aged 21 and older. We removed observations with missing values in candidate variables, including age, gender, pulse (beats/min), respiration (times/min), peripheral capillary oxygen saturation ($SpO_2$; %), diastolic blood pressure (mm Hg), systolic blood pressure (mm Hg), and comorbidities such as myocardial infarction, congestive heart failure, stroke, dementia, chronic pulmonary disease, peptic ulcer disease and kidney disease. For the SGH dataset, we obtained a sub-cohort with a total sample size of 81,110 by filtering the original SGH dataset for ED admissions in 2019. We focused on Chinese patients aged 21 and older while eliminating observations with missing values for the same candidate variables. In both datasets, the binary outcome of interest was inpatient mortality.

We designed five FL settings based on MIMIC and SGH data as summarized in Table 1. Settings A and B involved independent homogeneous partitioning of MIMIC data and SGH data. Settings C and D entailed independent heterogeneous partitioning of MIMIC data and SGH data based on age. Setting E conducted FL between the full cohorts of MIMIC and SGH data. The sample sizes for each client in descending order across these settings are available in Table 1.

**Experiments**

The GLORE framework was implemented by adapting the source code available at https://github.com/x1jiang/glore. To implement the FedAvg, FedAvgM, and $q$-FedAvg algorithms, we utilized the Flower[38] framework. As for the FedProx algorithm, we employed the source code from https://github.com/litian96/FedProx. For GLORE, the rounds of iterations were predetermined based on the distance between two consecutive $\beta$ values, rather than being determined by user input; additional information is provided in Supplementary A for reference. Conversely, for the four engineering-based frameworks, we determined the rounds of iterations through empirical testing and fine-tuning to identify suitable values for achieving convergence. We also observed that the choice of learning rates had minimal impact on the successful convergence of all engineering-based FL algorithms in the experiments conducted for this study, but only on the degree of time efficiency, with further details available in eTable 7 section D of the Supplementary Material.

## AUTHOR CONTRIBUTIONS

**Siqi Li:** Conceptualization, Data curation, Software, Formal analysis, Investigation, Methodology, Project administration, Writing – original draft, Writing – review & editing. **Di Miao:** Software, Formal analysis, Investigation, Methodology, Writing – original draft, Writing – review & editing. **Qiming Wu:** Software, Formal analysis, Investigation, Methodology, Writing – review & editing. **Chuan Hong:** Methodology, Validation, Investigation, Writing – review & editing. **Danny D'Agostino:** Methodology, Validation, Investigation, Writing – review & editing. **Xin Li:** Investigation, Writing – original draft, Writing – review & editing. **Yilin Ning:** Validation, Writing – original, Writing – review & editing. **Yuqing Shang:** Validation, Investigation, Writing – review & editing. **Huazhu Fu:** Validation, Investigation, Writing – review & editing. **Marcus Eng Hock Ong:** Validation, Investigation, Writing – review & editing. **Hamed Haddadi:** Validation, Investigation, Writing – review & editing. **Nan Liu:** Conceptualization, Investigation, Methodology, Project administration, Funding acquisition, Resources, Supervision.

## DATA AVAILABILITY

The simulations in this study can be reproduced using the source code available at https://github.com/nliulab/FL-Benchmark. The MIMIC cohort utilized in this study can be obtained from the master dataset generated with the pipelines outlined in Xie et al[37] for processing the original MIMIC-IV-ED data (https://physionet.org/content/mimic-iv-ed/1.0/), and then by following the details in Result section. The SGH data is confidential and not available to the public due to third-party data sharing restrictions.

## CODE AVAILABILITY

The code used for conducting this study is available at https://github.com/nliulab/FL-Benchmark.

## FUNDING

This work was supported by the Duke/Duke-NUS Collaboration grant. The funder of the study had no role in study design, data collection, data analysis, data interpretation, or writing of the report.

## COMPETING INTERESTS

NL, SL and MEHO hold a patent related to the federated scoring system. The other authors declare no competing interests.

## SUPPLEMENTARY INFORMATION

https://github.com/nliulab/FL-Benchmark/blob/main/Supplementary/FLB_algorithms_Supplement.pdf

**Table 1**. Summary of experimental settings.

**Table 2**. Performance assessment of real-world data prediction tasks, measured by AUC values.

**Table 3**. Overview of benchmark FL frameworks.

**Figure 1**. Comparison of prediction performance with a relatively small sample size and mean shifting.

**Figure 2**. Comparisons of estimated coefficients with a relatively small sample size and mean shifting.

**Figure 3**. Flowchart illustrating the selection criteria for applying FL algorithms to clinical structured data.

**Table 1.** Summary of experimental settings.

| Experiment Type | Setting | Details |
|---|---|---|
| Simulated data | I | Covariate shifts in mean ($\alpha = 0.1, 0.2, 0.3, 0.4$) |
| | II | Covariate shifts in standard deviation (($\alpha = 0.1, 0.2, 0.3, 0.4$) |
| | III | Effect size shifts (($\alpha = 0.1, 0.2$) |
| Real data | A | Artificial partitioned MIMIC data (homogeneously) with sample size (in decreasing order): 4536, 2993 and 1542 |
| | B | Artificial partitioned SGH data (homogeneously), with sample size (in decreasing order): 40,555, 26,766 and 13,789 |
| | C | Artificial partitioned MIMIC data (heterogeneously by age) with sample size (in decreasing order): 4085, 3288 and 1698 |
| | D | Artificial partitioned SGH data (heterogeneously by age) with sample size (in decreasing order): 36,434, 32,248 and 12,428 |
| | E | MIMIC data and SGH data with sample size 9071 (MIMIC) and 81,110 (SGH) |

**Table 2.** Performance assessment of real-world data prediction tasks, measured by AUC values.

(a) Prediction performance of federation settings among homogenously and heterogeneously partitioned MIMIC and SGH data.

| Testing Data | | | Model | | | | | | | | |
|---|---|---|---|---|---|---|---|---|---|---|---|
| | | | Central | Site 1 Local | Site 2 Local | Site 3 Local | GLORE | FedAvg | FedAvgM | q-FedAvg | FedProx |
| MIMIC | Homogeneous | Site1 | 0.773 | 0.705 | 0.709 | 0.788 | 0.773 | 0.801 | 0.801 | 0.797 | 0.780 |
| | | Site2 | 0.771 | 0.703 | 0.695 | 0.791 | 0.771 | 0.800 | 0.800 | 0.796 | 0.779 |
| | | Site3 | 0.774 | 0.716 | 0.730 | 0.780 | 0.774 | 0.798 | 0.799 | 0.793 | 0.778 |
| | | Average | 0.773 | 0.708 | 0.711 | 0.786 | 0.773 | 0.800 | 0.800 | 0.795 | 0.779 |
| | Heterogeneous | Site1 | 0.793 | 0.674 | 0.794 | 0.781 | 0.793 | 0.806 | 0.807 | 0.806 | 0.796 |
| | | Site2 | 0.785 | 0.686 | 0.783 | 0.774 | 0.785 | 0.804 | 0.804 | 0.802 | 0.788 |
| | | Site3 | 0.788 | 0.703 | 0.790 | 0.770 | 0.788 | 0.807 | 0.806 | 0.804 | 0.790 |
| | | Average | 0.789 | 0.688 | 0.789 | 0.775 | 0.789 | 0.806 | 0.806 | 0.804 | 0.791 |
| SGH | Homogeneous | Site1 | 0.862 | 0.858 | 0.860 | 0.862 | 0.862 | 0.864 | 0.864 | 0.858 | 0.862 |
| | | Site2 | 0.866 | 0.861 | 0.864 | 0.866 | 0.866 | 0.868 | 0.868 | 0.862 | 0.866 |
| | | Site3 | 0.866 | 0.861 | 0.865 | 0.865 | 0.866 | 0.869 | 0.869 | 0.863 | 0.866 |
| | | Average | 0.865 | 0.860 | 0.863 | 0.864 | 0.865 | 0.867 | 0.867 | 0.861 | 0.865 |
| | Heterogeneous | Site1 | 0.862 | 0.855 | 0.863 | 0.860 | 0.862 | 0.863 | 0.863 | 0.857 | 0.862 |
| | | Site2 | 0.865 | 0.859 | 0.866 | 0.863 | 0.865 | 0.866 | 0.866 | 0.860 | 0.865 |
| | | Site3 | 0.865 | 0.859 | 0.867 | 0.863 | 0.865 | 0.867 | 0.867 | 0.861 | 0.865 |
| | | Average | 0.864 | 0.858 | 0.865 | 0.862 | 0.864 | 0.865 | 0.865 | 0.859 | 0.864 |

(b) Prediction performance of federation between MIMIC and SGH data.

| Testing Data | Model | | | | | | | |
|---|---|---|---|---|---|---|---|---|
| | Central | MIMIC Local | SGH Local | GLORE | FedAvg | FedAvgM | q-FedAvg | FedProx |
| MIMIC | 0.789 | 0.797 | 0.785 | 0.789 | 0.792 | 0.792 | 0.811 | 0.787 |
| SGH | 0.850 | 0.833 | 0.848 | 0.850 | 0.852 | 0.852 | 0.842 | 0.850 |
| Average | 0.820 | 0.815 | 0.817 | 0.820 | 0.822 | 0.822 | 0.827 | 0.819 |

**Table 3**. Overview of benchmark FL frameworks.

| Framework | Description |
| --- | --- |
| GLORE[17] | A model-specific FL algorithm designed for logistic regression, capable of estimating coefficients, variance-covariance matrix and goodness-of-fit test statistics |
| FedAvg[7] | The pioneering model-agnostic FL algorithm, where the term 'FL' was originally coined |
| FedAvgM[18] | Built upon FedAvg and enhanced with server momentum[18] to handle non-identical data[18] |
| $q$-FedAvg[19] | Built upon FedAvg and enhanced with a novel optimization objective[19] to achieve a more uniform FL model performance across all clients[19] |
| FedProx[20] | Built upon FedAvg and enhanced to better handle both systems and statistical heterogeneity[7,28] in FL[20] |

**Figure 1.** Comparison of prediction performance with a relatively small sample size and mean shifting.

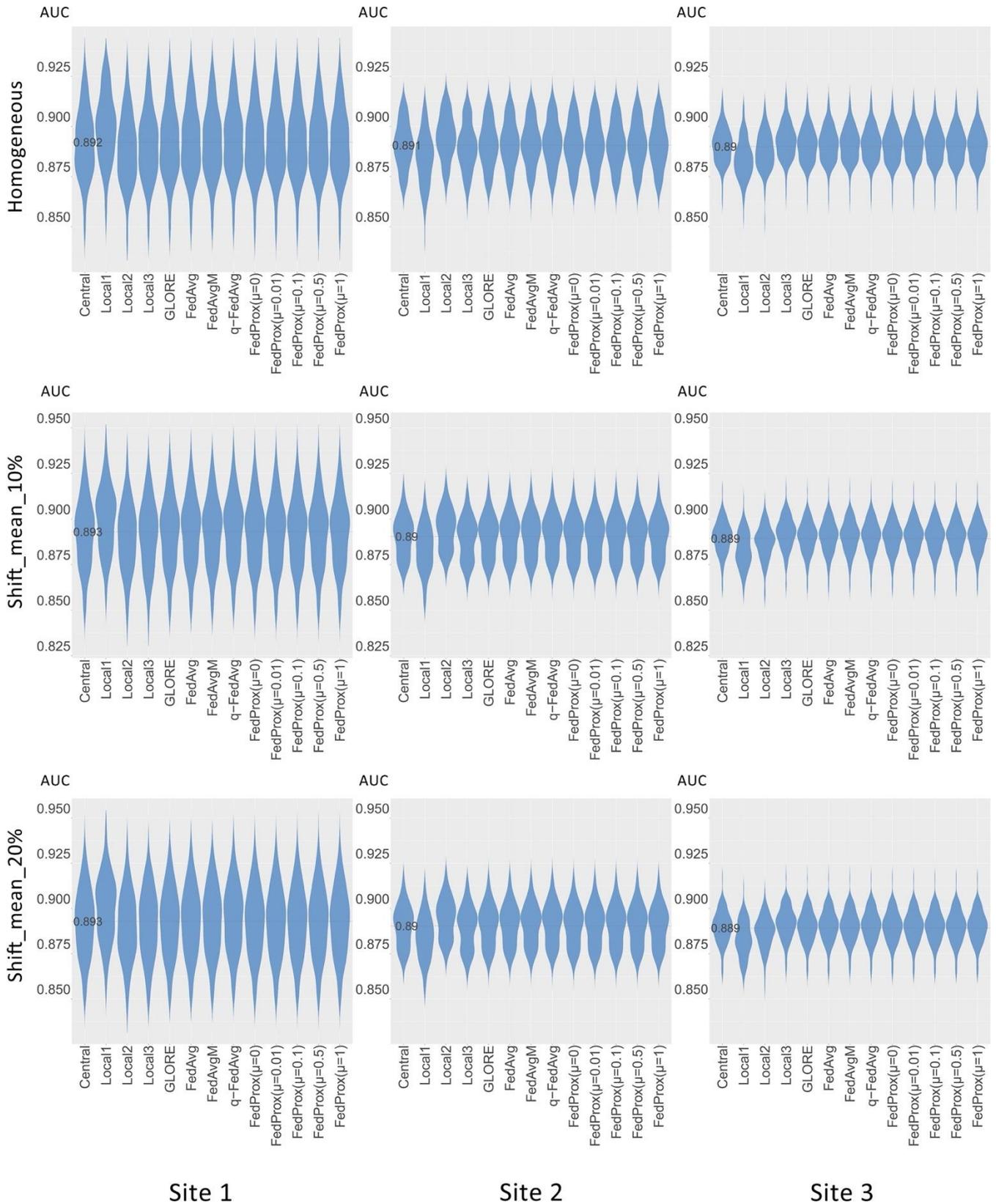

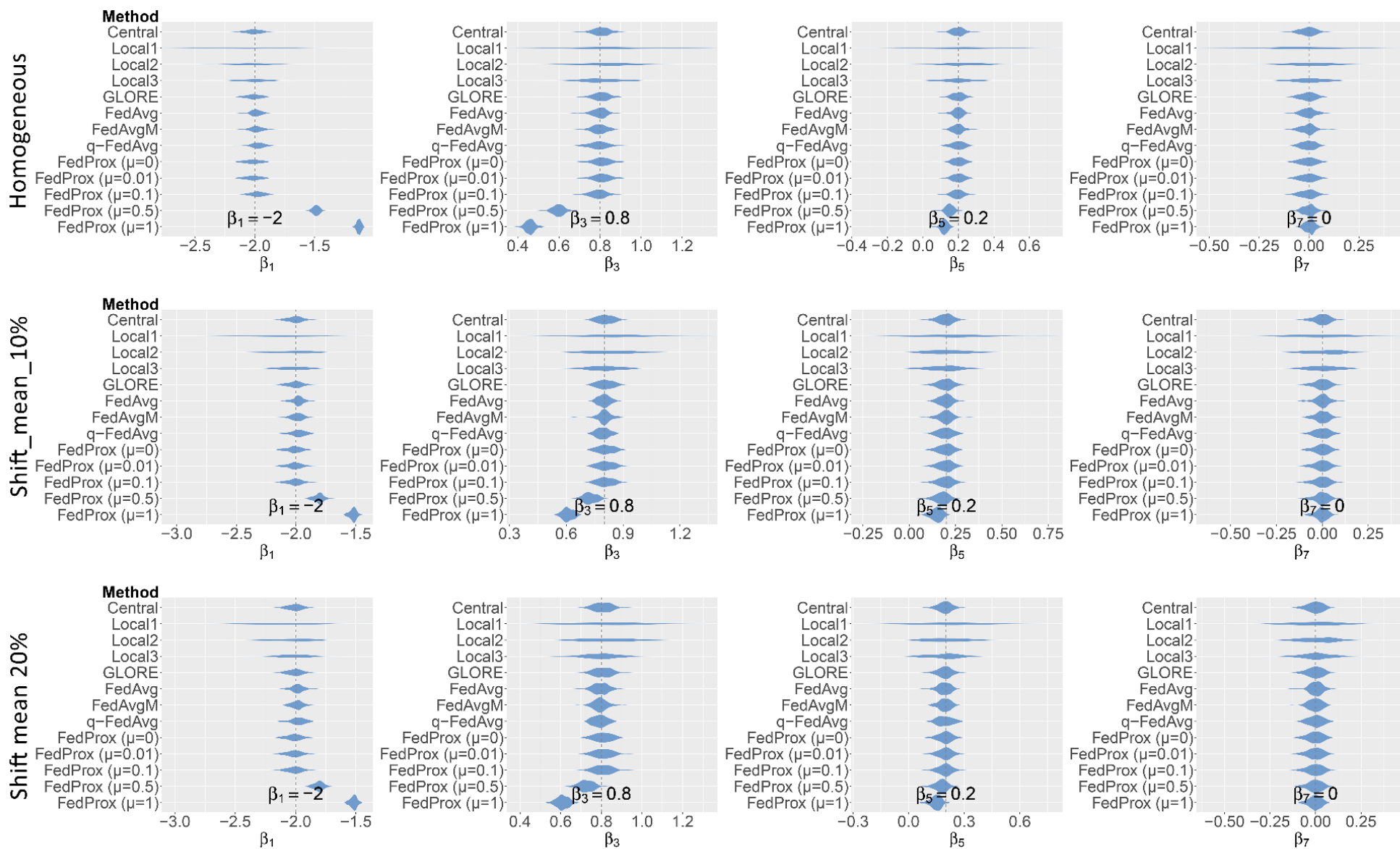

**Figure 2**. Comparisons of estimated coefficients with a relatively small sample size and mean shifting.

**Figure 3.** Flowchart illustrating the selection criteria for applying FL algorithms to clinical structured data.

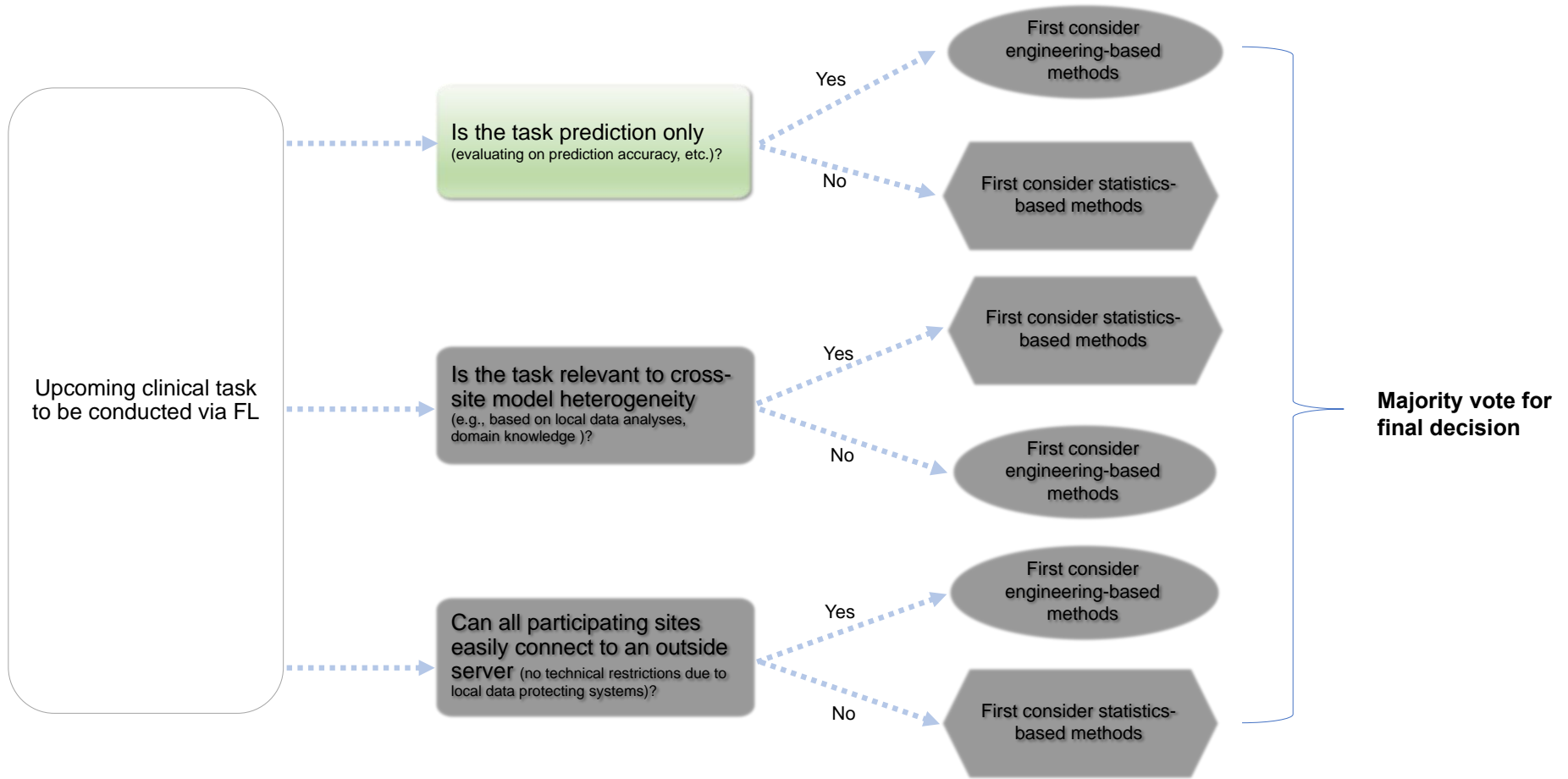